\documentclass[journal]{IEEEtran}
\usepackage{cite}
\usepackage{color,xcolor}
\usepackage{amsmath,amssymb,amsfonts}
\usepackage{algorithm}
\usepackage{algpseudocode}
\usepackage{graphicx}
\usepackage{textcomp}
\usepackage{wrapfig}
\usepackage{epstopdf}
\usepackage{textcomp}
\usepackage{array}
\usepackage{stfloats}
\usepackage{url}
\usepackage{booktabs}
\usepackage{threeparttable}
\usepackage{multirow}
\usepackage{bm}
\usepackage{gensymb}
\usepackage{graphics} 
\usepackage{epsfig} 
\usepackage{times} 
\usepackage{balance}
\usepackage{stfloats}
\usepackage{setspace}
\usepackage{multicol}

\begin{document}
	\title{LiDAR SLAMMOT based on Confidence-guided Data Association}
\author{Susu~Fang$^{2}$, and Hao~Li$^{*1,2}$ 
	\thanks{$^{1}$École d'Ingénieurs SJTU-ParisTech (SPEIT), Shanghai, 200240, China.}
	\thanks{$^{2}$Department of Automation, Shanghai Jiao Tong University (SJTU), Shanghai, 200240, China.}
	\thanks{$^{*}$Corresponding author: Hao Li (email: haoli@sjtu.edu.cn).}
}

	\maketitle
\begin{abstract}
In the field of autonomous driving or robotics, simultaneous localization and mapping (SLAM) and multi-object tracking (MOT) are two fundamental problems and are generally applied separately. Solutions to SLAM and MOT usually rely on certain assumptions, such as the static environment assumption for SLAM and the accurate ego-vehicle pose assumption for MOT. But in complex dynamic environments, it is difficult or even impossible to meet these assumptions. Therefore, the SLAMMOT, i.e., simultaneous localization, mapping, and moving object tracking, integrated system of SLAM and object tracking, has emerged for autonomous vehicles in dynamic environments. However, many conventional SLAMMOT solutions directly perform data association on the predictions and detections for object tracking, but ignore their quality. In practice, inaccurate predictions caused by continuous multi-frame missed detections in temporary occlusion scenarios, may degrade the performance of tracking, thereby affecting SLAMMOT. To address this challenge, this paper presents a LiDAR SLAMMOT based on confidence-guided data association (Conf SLAMMOT) method, which tightly couples the LiDAR SLAM and the confidence-guided data association based multi-object tracking into a graph optimization backend for estimating the state of the ego-vehicle and objects simultaneously. The confidence of prediction and detection are applied in the factor graph-based multi-object tracking for its data association, which not only avoids the performance degradation caused by incorrect initial assignments in some filter-based methods but also handles issues such as continuous missed detection in tracking while also improving the overall performance of SLAMMOT. Various comparative experiments demonstrate the superior advantages of Conf SLAMMOT, especially in scenes with some missed detections.
\end{abstract}

\section{Introduction}
\IEEEPARstart{S}{imultaneous} localization and mapping (SLAM) and multi-object tracking (MOT) are two fundamental problems for autonomous driving or robotic systems. SLAM aims to simultaneously model the surrounding static environment and accurately estimate the vehicle pose within the constructed map. MOT focuses on perceiving and estimating the states of surrounding moving objects. Currently, solutions for SLAM and MOT are usually applied separately and rely on certain assumptions. Most SLAM solutions rely on the assumption of a static environment. Many mature SLAM solutions have been proposed for single-vehicle applications \cite{shan2018lego, shan2020lio, xu2022fast}, but they still face significant challenges in dynamic environments containing a large number of moving objects, leading to degraded performance or even failure. On the other hand, multi-object tracking commonly assumes the accurate ego-vehicle pose as a known prior, placing little focus on its estimation. This dependence means that traditional MOT methods are significantly reliant on the precision of ego-pose estimation. However, in unknown dynamic environments, it is difficult to ensure the presence of reliable static structures, which often leads to an unreliable pose estimation.

Therefore, coupled SLAM and multi-object tracking can be a viable approach due to the inherent interdependence between the two challenges mentioned. Simultaneous localization, mapping, and moving object tracking (SLAMMOT), which was firstly presented in \cite{wang2007simultaneous}, coupling SLAM and MOT, allows for state estimation of ego-vehicle and moving objects simultaneously in dynamic environments. SLAMMOT can not only guarantee the performance of SLAM or MOT when the assumptions they rely on are difficult or even impossible to hold in complex dynamic environments, but also can mutually benefit each other. Specifically, for SLAM, considering the impact of moving objects transforms them from mere dynamic interference into auxiliary information that enhances pose estimation. And accurate ego-pose estimation from SLAM can significantly improve the estimation of object states for MOT.

In recent years, there has been significant progress in the research of SLAMMOT technology, such as vision-based SLAMMOT methods \cite{henein2020dynamic, bescos2021dynaslam, Li2024arXivVisualSLAMMOT} and LiDAR-based methods \cite{lin2023asynchronous,tian2024dl, ying2024imm}. While vision-based methods play an important role in the robotics and intelligent vehicle field, the advantages of LiDAR, such as adaptability to all kinds of weather conditions, high precision in distance measurements, high ranging accuracy, and wide horizontal field of view, have attracted more attention of researchers. However, there are still challenges in existing LiDAR-based SLAMMOT solutions. Most existing SLAMMOT methods directly adopt some kind of association strategy for data association between object prediction results and detections. However, object predictions are not always accurate, especially for objects that are missed in consecutive frames due to temporary occluded or distant objects, whose prediction errors may accumulate without receiving state updates. At the same time, some low-quality detections can also affect the results of data association. The inaccurate prediction and low-quality detection can destabilize the state of the matched track, or eventually lead to track failure and degrade the performance of SLAMMOT simultaneously. This tracking issue could pose a threat to driving safety if severe. Therefore, the confidence of predictions and detections should be considered during the data association in tracking of SLAMMOT.

\begin{figure}[t]
	\centering
	\includegraphics[scale=0.57]{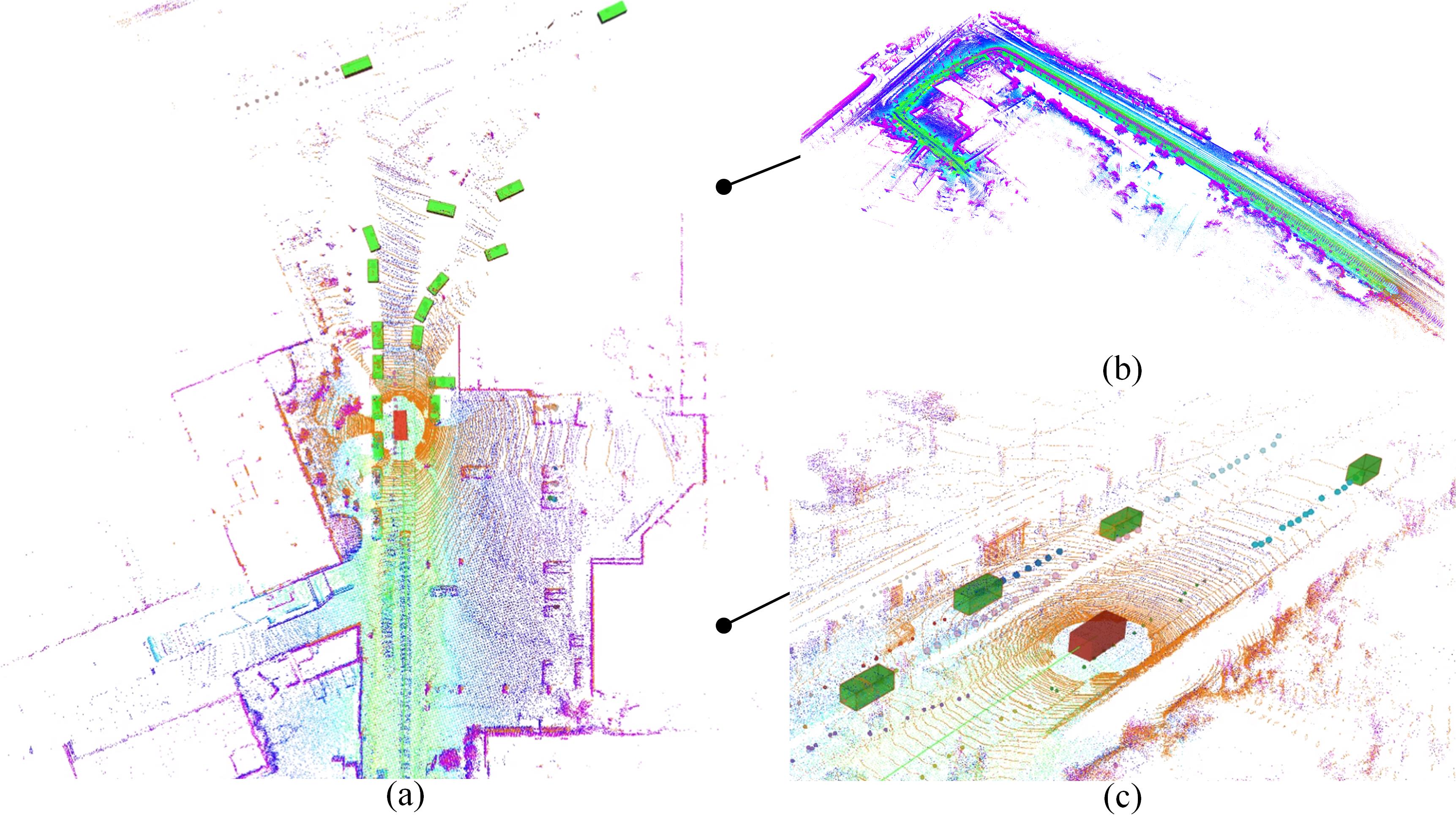}	
	\caption{Results of proposed Conf SLAMMOT solution in KITTI Tracking dataset \cite{geiger2012we}. (b) and (c) denote the generated complete point cloud map and details of the result shown in (a), respectively. Red and green fully covered bounding boxes denote the ego-vehicle and tracked objects, and red rectangular border denotes detected objects. The green solid line and the dotted lines represent the trajectories of ego-vehicle and tracked objects, respectively.}
	\label{Fig.1}
\end{figure}
In this paper, we present the Conf SLAMMOT, a tightly-coupled LiDAR-based SLAMMOT solution that performs confidence-guided data association in multi-object tracking, which can not only estimate the state of the ego-vehicle and moving objects simultaneously in dynamic environments, but also overcome the continuous missed detections caused by temporary occlusion or objects being far away in tracking. Fig. \ref{Fig.1} illustrates the results of the proposed Conf SLAMMOT solution in the KITTI Tracking dataset \cite{geiger2012we}. The Conf SLAMMOT includes a SLAM module, an object detection module, a confidence-guided data association based multi-object tracking relying on a factor graph, and a joint graph optimization backend for coupling and tracking. Herein, the SLAM and object detection modules are alternative and can be any representative solutions in the corresponding fields. The factor graph-based tracking utilizes the confidence of prediction and detection for its implicit data association, which can not only avoid the performance degradation caused by incorrect initial assignments, which is common in some filter-based methods, but also handle issues such as continuous missed detection in tracking. The partial factor graphs provided separately by the SLAM and object tracking modules are integrated into the graph optimization backend of Conf SLAMMOT, and the states of the ego-vehicle and the objects are jointly optimized. And the proposed method has achieved superior results in both SLAM and multi-object tracking tasks in various scenarios. The main highlights of the work are as follows:

$\bullet$ We present a LiDAR based SLAMMOT method that tightly couples SLAM and confidence-guided data association based multi-object tracking, allowing simultaneous estimation of the ego-vehicle and moving objects states, and achieving reliable tracking, especially in missed detection scenarios.

$\bullet$ We perform a predictor that applies a constant turn rate and velocity model to estimate the future state of object, and utilizes a prediction confidence model and the confidence score of detection to guide data association by flexibly adjusting the implicit search range of data association.

$\bullet$ The factor graph based tracking utilizes the confidence of prediction and detection for its data association, avoiding issues from wrong initial assignments in some filter-based methods and effectively recovering tracking of continuously missed detections due to objects being occluded or distant.
\section{Related Work}
\subsection{LiDAR based SLAM}
In the field of LiDAR-based SLAM, many solutions have been proposed. One of the most popular methods is LOAM \cite{zhang2017low}, which is a low-drift and real-time LiDAR odometry and mapping method. However, the performance of LOAM is limited when resources are scarce or when operating in noisy environments. To address these limitations, LeGO-LOAM \cite{shan2018lego} has been proposed, which is a lightweight and ground-optimized LOAM method and offers improved performance with limited resources. Additionally, auxiliary sensors like IMU can be closely integrated with LiDAR to improve SLAM performance. Researchers presents LIOM \cite{ye2019tightly}, which uses graph-based optimization to refine the extrinsic parameters of LiDAR and IMU, as well as IMU bias and robot pose. Following this, the LIO-SAM method \cite{shan2020lio} was introduced, estimating IMU bias within a factor graph framework and providing better real-time performance and more accurate state estimation compared to LIOM. To simplify computations, FAST-LIO \cite{xu2021fast} was developed, employing a tightly-coupled LiDAR-inertial odometry method based on an iterated Kalman filter with a new Kalman gain. Researchers then enhanced FAST-LIO with FAST-LIO2 \cite{xu2022fast}, which directly matches raw point clouds without feature extraction and boosts mapping velocity and odometry accuracy using iKD-Tree. Nevertheless, all of these LiDAR-based methods rely on the static environment assumption and are easily affected by dynamic objects.

For LiDAR-based SLAM in dynamic environments, researchers have also made some efforts. In \cite{sun2018motion} and \cite{zhao2019robust}, the researchers use convolutional neural networks (CNN) to detect moving objects from LiDAR point clouds, and simply filter all points belonging to dynamic objects for reliable SLAM in dynamic environments. In \cite{vaquero2019improving}, a two-stream CNN method was applied to segment potential dynamic objects from point clouds. After removing the points associated with the potential dynamic objects, the point clouds were used to perform SLAM. In \cite{qian2022rf}, researchers present the RF-LIO, which detects moving points in a LiDAR scan using adaptive multi-resolution range images and removes them, then realizes a tightly-coupled LiDAR inertial odometry in high dynamic environments. However, these methods may overlook valuable information, like points from stationary objects (e.g., parked vehicles), or face tracking errors due to the absence of a relationship between the ego-vehicle state and object state.
\subsection{Multi-object Tracking}
Multi-object tracking is a crucial module of the vehicle perception system. Existing tracking methods can be categorized into two groups: non-detection based methods and detection based tracking. In the non-detection based method, one approach is to integrate detection and tracking into a single network, as demonstrated by JDE \cite{wang2020towards} and CenterTrack \cite{zhou2020tracking}. Another approach is attention mechanism based tracking, exemplified by TrackFormer \cite{meinhardt2022trackformer}. However, these strategies suffer from drawbacks such as poor interpretability and generalization, and are primarily used in image based object tracking. Therefore, LiDAR based multi-object tracking typically employs a detection based method. 

In \cite{weng20203d}, researchers proposed AB3DMOT, which is based on a baseline for 3D multi-object tracking, featuring a lightweight architecture and low computational demands, enabling real-time operation. In \cite{wu20213d}, researchers introduced a new 3D multi-object tracker guided by prediction confidence for data association, which can better leverage object features for multi-object tracking in point clouds. These two methods are both advanced LiDAR-only real-time 3D multi-object tracking techniques, and they utilize detection results obtained from LiDAR point cloud data and employ Kalman filtering with Hungarian matching and greedy algorithm respectively for state estimation and data association. However, if the initial assignment in data association is incorrect, such filter-based approaches often struggle, which can easily occur. In \cite{poschmann2020factor}, researchers present a factor graph-based 3D multi-object tracking (FG-3DMOT) method, which represents the detection results as a Gaussian mixture model (GMM) and is integrated into a factor graph framework. FG-3DMOT does not rely on explicit and fixed assignments; instead, it flexibly assigns all detections to all objects simultaneously. The assignment issue is addressed implicitly and jointly with object state estimation by nonlinear least squares optimization. However, such methods ignore the consideration of the quality of detection and object prediction. For example, in cases of continuous missed detections, the predicted states can accumulate errors leading to inaccuracy, thereby affecting tracking performance. Additionally, there are some camera-LiDAR fusion-based multi-object tracking methods. In \cite{wang2022deepfusionmot}, researchers present a 3D multi-object tracking framework based on camera-LiDAR fusion with deep association, achieving seamless integration of 2D and 3D trajectories for more accurate and robust tracking, especially when tracked objects are far away. And in \cite{wang2023camo}, researchers present CAMO-MOT, which introduces a combined appearance-motion optimization technique that significantly reduces tracking failures caused by occlusions and false detections by using camera and LiDAR.
\subsection{Coupled SLAM and Object Tracking}
Currently, there have been many research efforts in coupled SLAM and object tracking. Specifically, researchers perform separate SLAM and object tracking systems simultaneously and estimate the ego-vehicle state and object state with a loosely-coupled approach \cite{ma2024semantic}. In \cite{lim2015monocular}, visual odometry and persistent tracking are performed simultaneously to detect humans in dynamic environments. Additionally, in \cite{sun2018motion}, researchers use neural networks to estimate object motion or segment potential dynamic objects from LiDAR point clouds, and SLAM is performed after filtering out dynamic object points. Another category integrates vehicle state estimation and object tracking into a unified optimization problem for tightly coupled SLAM with object tracking, which mainly revolve around visual-based and range sensor-based methods. In \cite{lin2010stereo}, researchers propose a tightly-coupled approach for SLAMMOT based on stereo visual information. In \cite{henein2020dynamic}, Dynamic SLAM is proposed, which performs object detection and tracking based on deep learning during the operation of SLAM in a dynamic environments. Furthermore, DynaSLAM II \cite{bescos2021dynaslam} demonstrates tightly-coupled multi-object tracking and visual SLAM through instance semantic segmentation and ORB features.

Regarding range sensor-based solutions, researchers in \cite{wang2007simultaneous} propose a feasible tightly-coupled SLAMMOT system based on a laser scanner, deriving a dynamic Bayesian networks algorithm for coupling. In \cite{vu2011grid}, researchers present an online SLAMMOT method using a grid-based approach, achieving pose estimation and object tracking in two-dimensional space. In \cite{tian2024dl}, researchers present DL-SLOT, which uses sliding window-based graph optimization to simultaneously optimize the vehicle state and dynamic-static object states based on LiDAR. Based on DL-SLOT, the researchers present LIMOT \cite{zhu2024limot}, which is also a tightly-coupled multi-object tracking and LiDAR-inertial odometry system, and it presents a dynamic feature filtering method for the LiDAR odometry module. Although their introduced trajectory-based state prediction improves the tracking performance, this sliding window-based local optimization reduces its ability to perform global optimization for ego-vehicle and object states. In \cite{ying2024imm}, researchers propose the IMM-SLAMMOT, which integrates interactive multiple model into object tracking for SLAMMOT to deal with the ambiguous motion patterns of objects. The data association in all above tracking methods follows fixed assignments, which tend to struggle if the initial assignment is incorrect. In \cite{lin2023asynchronous}, researchers propose LIO-SEGMOT for asynchronous estimation of simultaneous ego-motion state and multiple object tracking, which integrates the factor graph based tracking that does not rely on explicit and fixed assignments in data association. However, these methods ignore considering the confidence during tracking, especially in scenarios with continuous missed detection due to occluded or distant objects, which can lead to performance degradation, affecting the accuracy and stability of SLAMMOT.
\section{THE PROPOSED APPROACH}
\subsection{Overview of the Proposed Framework} 
\begin{figure}[htbp]
	\centering
	\includegraphics[scale=0.77]{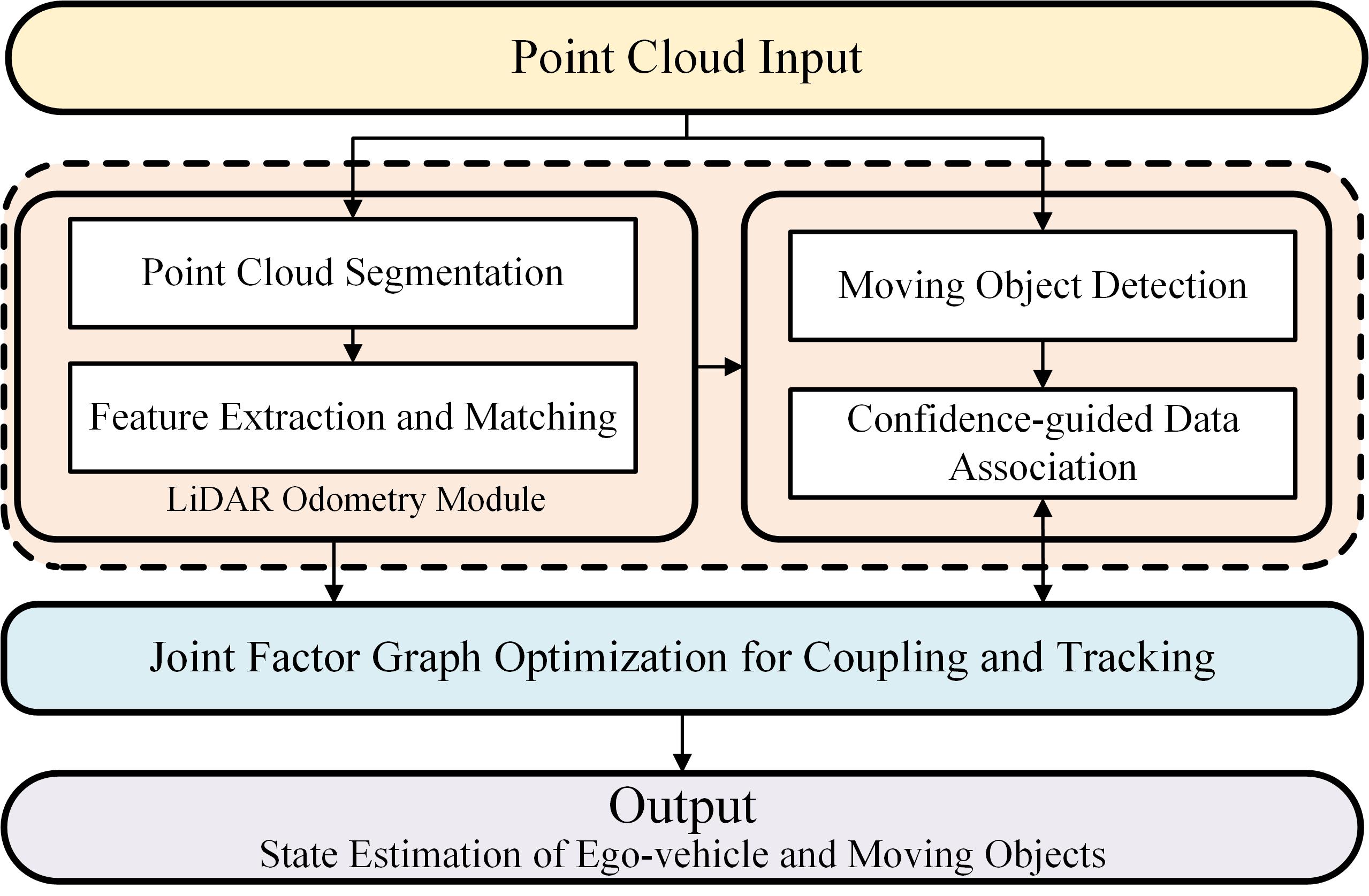}	
	\caption{Overall architecture of the presented Conf SLAMMOT solution.}
	\label{SLAMMOT2}
\end{figure}

The proposed Conf SLAMMOT solution extends from the foundation of LiDAR SLAM based on graph optimization. The architecture of the Conf SLAMMOT solution is shown in Fig. \ref{SLAMMOT2}, which mainly consists of an alternative LiDAR odometry module, an alternative moving object detection module, and a joint graph optimization backend, which is a tightly-coupled LiDAR SLAM and multi-object tracking method in dynamic environments for simultaneously estimating the states of the ego-vehicle and moving objects. The 3D point cloud information serves as the input for the LiDAR odometry module and the moving object detection network, respectively. The LiDAR SLAM method (the module used here is from LeGO-LOAM \cite{shan2018lego}), as shown in Sec. III-B, consists of a replaceable LiDAR odometry module and partial graph optimization for the joint optimization backend of the overall system. For object detection, the advanced deep learning-based method, PV-RCNN \cite{shi2020pv}, is used, as shown in Sec. III-C. As for the object tracking part of the system, inspired by FG-3DMOT \cite{poschmann2020factor}, we use a max-mixture model for implicitly associating object prediction states with detection results and integrate it with a prediction confidence model for confidence-guided data association. To demonstrate the improvements we have made in data association, we display this implicit data association as a module in the figure. It should be noted that this implicit data association is managed as part of the factor graph optimization and is to be jointly optimized with the overall system, as shown in Sec. III-D and E. Finally, the joint factor graph optimization backend outputs the state estimation of the ego-vehicle and moving objects simultaneously, as shown in Sec. III-F.
\subsection{LiDAR Odometry Module}
In presented Conf SLAMMOT solutions, the SLAM can be any representative LiDAR SLAM framework. Due to the several sub-modules of the proposed system, the lightweight of each module should be considered under conditions of limited computing resources. Also, the IMU is not a mandatory sensor in many cases. Therefore, LeGO-LOAM \cite{shan2018lego} is applied for LiDAR odometry module, which maintains high localization and mapping accuracy while consuming fewer computing resources compared to the LOAM \cite{zhang2017low}.

The LeGO-LOAM receives input from a 3D LiDAR and outputs a 6 Degrees of Freedom (DoF) ego-pose estimate and a 3D point cloud map. This framework is divided into five submodules: segmentation, feature extraction, LiDAR odometry, LiDAR mapping, and trajectory integration \cite{shan2018lego}.The LeGO-LOAM project the raw point cloud into a distance image and differentiating between ground points and segmented points before feature extraction. It then matches feature points of the same category and incorporates the iSAM2 \cite{kaess2012isam2} optimizer in the backend. Finally, the ego-vehicle pose estimation obtained by this module will provide to the multi-object tracking and joint graph optimization backend.
\subsection{Object Detection Module}
For object detection in Conf SLAMMOT, it can be the any representative detection method. Herein, we use the PV-RCNN, which is an advanced framework capable of achieving precise 3D object detection from point clouds  with manageable memory consumption. PV-RCNN integrates the advantages of 3D voxel CNN with sparse convolution \cite{yan2018second} and PointNet-based networks \cite{qi2017pointnet++} to learn more discriminative point cloud features. The method utilizes a 3D voxel sparse CNN as the backbone network and employs a voxel set abstraction module to convert 3D scenes into keypoints, followed by a keypoint-to-grid Region of Interest (RoI) feature abstraction for refinement, used for proposal confidence prediction and location refinement. The two-step strategy of this framework effectively combines voxel-based and point-based learning, demonstrating accuracy and efficiency in 3D object detection using point cloud data without additional sensor inputs. PV-RCNN's innovative approach and its performance on the KITTI benchmark have established it as a popular solution in the field of autonomous driving object detection \cite{geiger2012we}. 
\subsection{Object State Prediction}
For object state prediction, considering prediction accuracy and efficiency, most current 3D object detection algorithms typically employ motion models to predict future object states, with constant velocity (CV)\cite{scheidegger2018mono,simon2019complexer,weng20203d}, constant acceleration (CA)\cite{wu20213d}, and constant turn rate and velocity (CTRV) models \cite{fang2024multi} being widely utilized. The CV model assumes that the object moves at a constant velocity, disregarding the effects of acceleration. This model is particularly effective when object motion is smooth, without sharp turns or changes in acceleration. The CA model assumes the object moves with constant acceleration, making it suitable for scenarios where the object's acceleration is relatively stable. The CTRV model assumes the object maintains a constant turn rate and velocity during turning, which is applicable to cases involving curved motion. Additionally, some learning-based tracking methods explore predicting states by learning object position transformations from historical trajectory data through neural networks \cite{kim2018multi,hu2019joint,farhodov2020lstm}; however, they are typically several times slower than the aforementioned motion models and often face challenges in practical applications. Therefore, taking into account both prediction accuracy and efficiency, as well as addressing the missed detection issue we discuss, where objects are more likely to be lost during turning scenarios, we adopt the CTRV model to predict the future states of objects.

After obtaining the object state predictions and detection results, most traditional methods directly associate them (explicitly or implicitly) to solve the assignment problem. However, the predicted states of objects are not always accurate; for instance, after missing detections in several consecutive frames, they can accumulate significant errors, thus enlarging the possible range of detected objects when resuming tracking. Meanwhile, the quality of detection results also needs to be considered. Therefore, we have incorporated a prediction confidence model \cite{wu20213d} into our present system to make data association more sensitive to the quality of each prediction.
\begin{equation}
\small
\hat c_{pre,k}^{i}=c_{pre,k}^{i}-\alpha c_{pre,k-1}^{i}
\end{equation}

Here, \begin{small}$c_{pre}\in (0,1]$\end{small} denotes the prediction confidence and \begin{small}$\alpha \in [0,1]$\end{small} denotes a parameter that adjusts the overall impact of prediction confidence on data association and is determined by dataset. Applying this model can address the issue of continuous multi-frame missed detections, where when a miss occurs, the prediction confidence decreases, and the implicit search range for data association is subsequently enlarged. This flexible adjustment of the implicit search range for data association can better recover re-tracking after continuous missed detection, preventing ID switches.
\subsection{Confidence-guided Implicit Data Association}
The designed implicit confidence-based data association solves the assignment problem between detection results and predicted states with their corresponding confidence. This method employs the GMM to simultaneously represent all detection results, and we utilize the max-mixture model \cite{olson2013inference,poschmann2020factor} to represent the full GMM of each object that is integrated with the prediction confidence model. Specifically, we apply the GMM as a probabilistic model of detection results that overcomes the limitations of simple single Gaussian models and breaks away from the least squares formulation inherent in the maximum likelihood problem, selecting one from all possible predicted states of \begin{small}${\bf{\bar{O}} }_{k}$\end{small} such that the likelihood of observing the current data given the detection results \begin{small}${\bf{Z}}_{k}$\end{small} is the highest. This confidence-guided implicit data association integrates the assignment problem into the factor optimization to jointly estimate the object state and solve the data association implicitly as part of the optimization, which allows for a flexible association between predicted states and detections that can change during the optimization process, unlike most online trackers, and goes beyond merely estimating the association by optimizing state positions across the entire sequence to account for inaccuracies in object prediction and detection. By applying the negative logarithm, the optimization problem arising from the maximum likelihood formulation can be reformulated as follows \cite{pfeifer2019expectation}:
\begin{equation}
\footnotesize
\begin{aligned}
{{\bf{\hat O}}} = \mathop {\arg \min }\limits_O \sum\limits_i { - \ln ({\bf{P}}({\bf{Z}}{}_{i,k}|{{\bf{\bar{O}}}}_{i,k}))}
\end{aligned}
\end{equation}

Where the optimal estimated \begin{small}${{\bf{\hat O}}}_{k}$\end{small} denotes the maximum-likelihood estimator of \begin{small}${ \bf{\bar{O}} }_{k}$\end{small}; \begin{small}${\bf{P}}({\bf{Z}}{}_{i,k}|{{\bf{\bar{O}}}}_{i,k})$\end{small} denotes the conditional probability that associates the $i$-th predicted object state \begin{small}${{\bf{\bar{O}}}}_{i,k}$\end{small} with the corresponding detection result \begin{small}${{\bf{Z}}_{i,k}}$\end{small} that belongs to it. For a sum of Gaussians with $n$ components ($c$), the conditional probability \begin{small}${\bf{P}}$\end{small} is defined as:
\begin{equation}
\footnotesize
\begin{aligned}
{\bf{P}}({\bf{Z}}{}_{i,k}|{{{\bf{\bar{O}}}}_{i,k}}) \propto \sum\limits_{j = 1}^n {{c_j}}  \cdot \exp ( - \frac{1}{2}{\left\| {\Sigma _j^{ - \frac{1}{2}}({{\bf{\bar{O}}}}_{i,k} - {\bf{T}}_{E,k}^{{M_E}} \cdot {\bf{Z}}{}_{j,k})} \right\|^2})
\end{aligned}
\end{equation}

Herein, $m$ deonots the sum of detection measurements; $j$ denotes the $j$th detection. \begin{small}${\bf{T}}_{E,k}^{{M_E}}$\end{small} denotes the transformation from ego-vehicle coordinate system to map coordinate system \begin{small}${M_E}$\end{small}. And \begin{small}${c_j} = {\omega _j} \cdot \det (\Sigma _j^{ - \frac{1}{2}})$\end{small}, where \begin{small}${\omega _j}$\end{small} and \begin{small}$\Sigma _j$\end{small} denotes the weight and uncertainty of $j$th detection. Herein, the \begin{small}$\Sigma _j$\end{small} considers the confidence score of detection results, denoting \begin{small}$\Sigma _j=\Gamma \cdot (1-c_{det,k}^j) \cdot \beta$\end{small}, instead of using the overall uncertainty of the detector \begin{small}$\Gamma$\end{small} determined by sensor’s error characteristic. \begin{small}$c_{det,k}$\end{small} denotes the confidence score of the corresponding detection box, and \begin{small}$\beta$\end{small} denotes a scaling parameter that can vary depending on the detector and the dataset. And for the conditional probability \begin{small}\(\mathbf{P}\)\end{small}, the logarithm cannot be pushed inside due to the summation, so we need to calculate the log-likelihood differently. There are two suggested solutions: an approximate max-mixture method \cite{olson2013inference} and an exact sum-mixture method \cite{pfingsthorn2013simultaneous}. And the max-mixture approach simplifies calculations compared to the sum-of-Gaussians method while maintaining similar effectiveness. Therefore, the summation of a GMM can be replaced by the maximum operator as following equation (4), and we push inside the logarithm for maximum becoming a minimum as shown in equation (5):
\begin{equation}
\footnotesize
\begin{aligned}
- \ln ({\bf{P}}) =  - ln\mathop {\max }\limits_j ({c_j} \cdot \exp ( - \frac{1}{2}{\left\| {\Sigma _j^{ - \frac{1}{2}}({{\bf{\bar{O}}}}_{i,k}- {\bf{T}}_{E,k}^{{M_E}} \cdot {\bf{Z}}{}_{j,k})} \right\|^2}))\\
\end{aligned}
\end{equation}
\begin{equation}
\footnotesize
\begin{aligned}
- \ln ({\bf{P}}) = \mathop {\min }\limits_j ( - ln{c_j} + \frac{1}{2}{\left\| {\Sigma _j^{ - \frac{1}{2}}({{\bf{\bar{O}}}}_{i,k} - {\bf{T}}_{E,k}^{{M_E}} \cdot {\bf{Z}}{}_{j,k})} \right\|^2})
\end{aligned}
\end{equation}

To maintain a uniform surface in the optimization formula, we perform dimensional separation between the logarithmic normalization term and the quadratic error term. This not only aids in achieving better convergence but also allows us to consider the quadratic error term individually since the log-normalization term is a constant. The least squares problem in this optimization is expressed as follows:
\begin{equation}
\footnotesize
\begin{aligned}
{\left\| {{{\bf{e}} _{i,k}}} \right\|^2} = \mathop {\min }\limits_j {\left\| \begin{array}{l}
	\sqrt { - 2 \cdot \ln \frac{{{c_j}}}{{{c_{\max }}}}} \\
	\Sigma _j^{ - \frac{1}{2}}({{\bf{\bar{O}}}}_{i,k} - {\bf{T}}_{E,k}^{{M_E}} \cdot {\bf{Z}}{}_{j,k})
	\end{array} \right\|^2}
\end{aligned}
\end{equation}

The term \begin{small}$- \ln \frac{{{c_j}}}{{{c_{\max }}}}$\end{small} represents the independent dimension of the vectorized error function, where \begin{small}$c_{\max }$\end{small} is a normalization constant to ensure that the expression under the square root is positive. \begin{small}${\bf{e}}$\end{small} denotes the weighted error between the $i$th object state and $j$th detection measurement. Then, considering the prediction confidence model in equation (1), the quadratic error term can be rewritten as follows:
\begin{equation}
\footnotesize
\begin{aligned}
\hat c_{{pre,k}}^{i} \cdot \mathop {\min }{\left\| 
	\Sigma _i^{ - \frac{1}{2}}({{\bf{\bar{O}}}}_{i,k} - {\bf{T}}_{E,k}^{{M_E}} \cdot {\bf{Z}}{}_{i,k})\right\|^2} < \sigma
\end{aligned}
\end{equation}

Herein, \begin{small}$\sigma > 0$\end{small} denotes an association threshold to determine whether the detection belongs to any existing object with prediction confidence. Specifically, for each predicted state, a maximum likelihood model is employed to correlate all detection results, and the minimum value of the aforementioned formula is selected. If this minimum value is less than the threshold, the detection is considered associated. Compared to approaches that do not take into account the prediction confidence, this method effectively enlarges the implicit range of association, which is highly effective for resuming tracking after consecutive missed detections, as shown in Fig. \ref{FGO}(b).

The aforementioned implicit data association is integrated into graph optimization, after which the tracking state for matched detection states and predicted states are updated by using the optimization, in order to obtain the updated object states. Concurrently, the prediction confidence, based on the confidence of the corresponding detection results \begin{small}$c_{det,k}$\end{small}, will also be updated as follows:
\begin{equation}
\footnotesize
c_{pre,k}^m = \left\{ \begin{array}{l}
1,\quad\quad\quad\quad \quad\quad\quad\;\; c_{pre,k - 1}^m = 0,c_{pre,k}^m \ne 0\\
\hat c_{pre,k}^m + \alpha c_{det,k}^m,\quad otherwise
\end{array} \right.
\end{equation}
where $m$ denotes the matched paris; \begin{small}$ c_{pre,k-1}^m$\end{small} denotes the confidence of missed detection at the moment \begin{small}$k-1$\end{small}.

For unmatched detection and prediction results, we employ the following tracking management strategy. Detection results with very low confidence are directly discarded, while those with higher confidence are initialized to establish new tracks. For unmatched predicted object states, there are two scenarios. The first scenario is when the object naturally disappears from the field of view. The second scenario occurs in challenging situations where the object is temporarily occluded by other objects or is too far from the sensor, leading to missed detections. We set a threshold \begin{small}$N$\end{small} for the number of prediction frames to distinguish between these two cases. If a predicted state cannot be updated by detection states for more than \begin{small}$N$\end{small} frames (set to 12 here), it is considered to have naturally vanished. Such predicted states are removed and no longer tracked. Otherwise, the predicted states are retained, as the object may have been temporarily missed and could reappear in future frames, as illustrated in Fig. \ref{FGO}(b), where the state performs prediction from time \begin{small}$k-t$ \end{small}to \begin{small}$k-1$\end{small} and re-associates data guided by confidence at time $k$ to resume tracking.
\subsection{Joint Factor Graph Optimization}
\begin{figure}[htbp]
	\centering
	\includegraphics[scale=0.60]{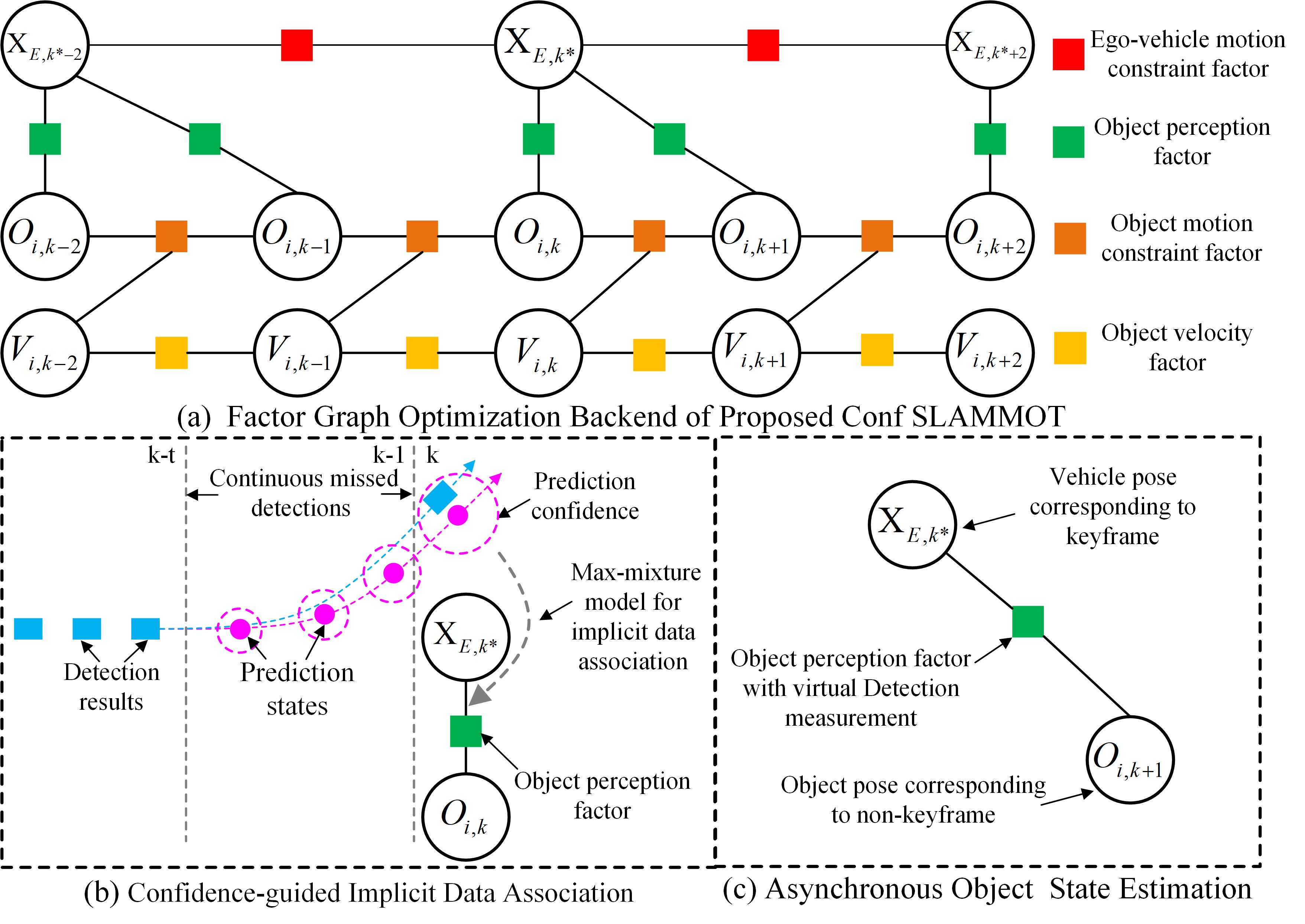}	
	\caption{The factor graph model in proposed Conf SLAMMOT solution. (a) Joint factor graph optimization backend for coupling and tracking. (b) is an explanation subfigure of confidence-guided implicit data association. (c) is an explanation subfigure of asynchronous object state estimation.}
	\label{FGO}
\end{figure}

The unified graph optimization backend couples SLAM with the multi-object tracking to jointly optimize the state estimation of the ego-vehicle and moving objects. It extends the factor graph optimization of the LiDAR SLAM module and is integrated with factor graph-based tracking. As depicted in Fig. \ref{FGO}(a), all variable nodes represent the estimated 3D states. The node corresponding to the ego-vehicle pose \begin{small}${{\bf{X}}_{E,k^*}}$\end{small} at the timestamp of selected key frames $k*$ is obtained through LiDAR odometry. The edge connecting two nodes represents the ego-vehicle motion constraint factor ${{\bf{e}} _{odo}}$:
\begin{equation}
\small
{{\bf{e}} _{odo}^{k^*-1,k^*}}({{\bf{X}}_{k^*-1}},{{\bf{X}}_{k^*}})={{\bf{T}}_{k^*-1}^{k^*}} {{\bf{X}}_{k^*-1}}-{{\bf{X}}_{k^*}}
\end{equation}

The object perception factor connects the ego-vehicle pose node and the $i$-th moving object pose node \begin{small}${{{\bf{O}}}_{i,k}}$\end{small} that deals with the implicit data association between prediction objects and detection measurements by the confidence-guided implicit data association introduced in Sec. III-D, as shown in Fig. \ref{FGO}(b). For computational efficiency in LiDAR odometry, keyframe selection is used for backend state estimation, meaning that the vehicle state is updated only in keyframes, and non-keyframe LiDAR data, including moving object detections, is ignored. Conventional methods typically estimate vehicle and object states synchronously, but factor graph optimization-based solutions do not have this limitation. Inspired by \cite{lin2023asynchronous}, we use asynchronous state estimation to utilize object information from non-keyframes. This involves transforming object detection results from non-keyframes to virtual moving object detection measurements in the latest keyframes, as shown in Fig. \ref{FGO}(c). This transformation uses the relative vehicle state estimates at timestamps $k$ and $k^*$ through scan matching. Such an object perception factor is denoted as:
\begin{equation}
\small
{{\bf{e}} _{op}^{k^*,k}} ({{\bf{X}}_{E,k^*}},{{{\bf{{O}}}}_{i,k}}|{{\bf{Z}}_{i,k}})={{{\bf{{O}}}}^{{M_{E}}}_{i,k}}-{{\bf{X}}^{{M_{E}}}_{E,k^*}}{{\bf{Z}}_{i,k}}
\end{equation}

The edge between two object pose nodes is denoted as the object motion constraint factor ${{\bf{e}} _{mov,i}^{k-1,k}}$ shown in equation (11), where $f(\cdot)$ denotes the CTRV movement function of object states driven by its velocity \begin{small}${{\bf{V}}_{i,k}}$\end{small} in a time step \begin{small}$\Delta T$\end{small}, where the \begin{small}${{\bf{V}}_{i,k}}=[{{\bf{v}}_{i,k}}, {{\bf{\omega}}_{i,k}}]^T$\end{small} denotes the linear velocity and angular velocity respectively. The edge constraints between two poses can be represented by velocity factors ${{\bf{e}} _{v,i}^{k-1,k}}$ in equation (12), which assume a reasonable approximation that the object moves with a constant velocity over a short time.
\begin{equation}
\footnotesize
{{\bf{e}} _{mov,i}^{k-1,k}}({{{\bf{{O}}}}_{i,k-1}},{{{\bf{{O}}}}_{i,k}},{{\bf{V}}_{i,k-1}})=f({{{\bf{{O}}}}_{i,k-1}},{{\bf{V}}_{i,k-1}},\Delta T)-{{{\bf{{O}}}}_{i,k}}
\end{equation}
\begin{equation}
\small
{{\bf{e}} _{v,i}^{k-1,k}}({{\bf{V}}_{i,k-1}},{{\bf{V}}_{i,k}})={{\bf{V}}_{i,k-1}}-{{\bf{V}}_{i,k}}
\end{equation}

Finally, the joint factor graph optimization is defined as minimizing the sum of nonlinear least-square errors:
\begin{equation}
\footnotesize
\begin{aligned}
{X^{*}}&=  \mathop {\arg \min }\limits_\chi \sum\limits_{k^*-1,k^*} {\left\| 	{{\bf{e}} _{odo}^{k^*-1,k^*}} \right\|_{{\sum _{odo}}}^2 }\!\!\!+ {\left\| {{{\bf{e}} _p}({\chi _0})} \right\|^2}\\
\!\!\!+&\sum\limits_{i,k-1\hfill\atop k^*,k\hfill}({\left\| {{\bf{e}} _{op}^{k^*,k}} \right\|_{{\sum_{op} }}^2}\!\!\!+{\left\| {{\bf{e}} _{mov,i}^{k-1,k}} \right\|_{{\sum _{mov}}}^2}\!\!\!+{\left\| 	{{\bf{e}} _{v,i}^{k-1,k}} \right\|_{{\sum _v}}^2})
\end{aligned}
\end{equation}

Where \begin{small}$\chi$\end{small} denotes the set of all variables, and \begin{small}${\left\| {{{\bf{e}} _p}({\chi _0})} \right\|^2}$\end{small} denotes the prior information error term. \begin{small}$\sum\nolimits_{{odo}}$, $\sum\nolimits_{{op}}$, $\sum\nolimits_{{mov}}$, $\sum\nolimits_{{v}}$\end{small}, and \begin{small}$\sum\nolimits_{{p}}$\end{small} denote the covariance or standard deviation matrix of each variable factor, respectively.
\section{Experimental Evaluation}
\subsection{Experiment Conditions}
We provide a comprehensive overview of the experiments conducted to evaluate the performance of the proposed Conf SLAMMOT and the baseline methods in various aspects. We conducted experiments on several sequences of the KITTI Tracking dataset\cite{geiger2012we}. This dataset was primarily collected on urban roads and highways, and it includes raw point clouds, IMU/GPS data, and ground truth labels that can be used for tracking evaluation. We use 18 sequences of dataset with ground truth label data as evaluation sequences, excluding sequences 16, 17, and 20, because they are either highway sequences that cause SLAM to fail or sequences that primarily focus on pedestrian objects. To highlight the coupling advantages of the Conf SLAMMOT over standalone LiDAR SLAM or multi-object tracking systems, we implement comparative experiments by several methods to evaluate our proposed solution for state estimation of ego-vehicle and objects. And we also evaluate the overall performance of Conf SLAMMOT with other baseline methods in different sequences. For the parameters and thresholds used in confidence-guided implicit data association, we set $\alpha$ to 0.03,  $\beta$ to 80 and $\sigma$ to 6.5 for the applied method and dataset. Additionally, we evaluate the time consumption of each module in the proposed solution.
\subsection{Ego-vehicle Pose Estimation Evaluation}
\begin{figure*}[htbp]
	\centering
	\includegraphics[scale=0.85]{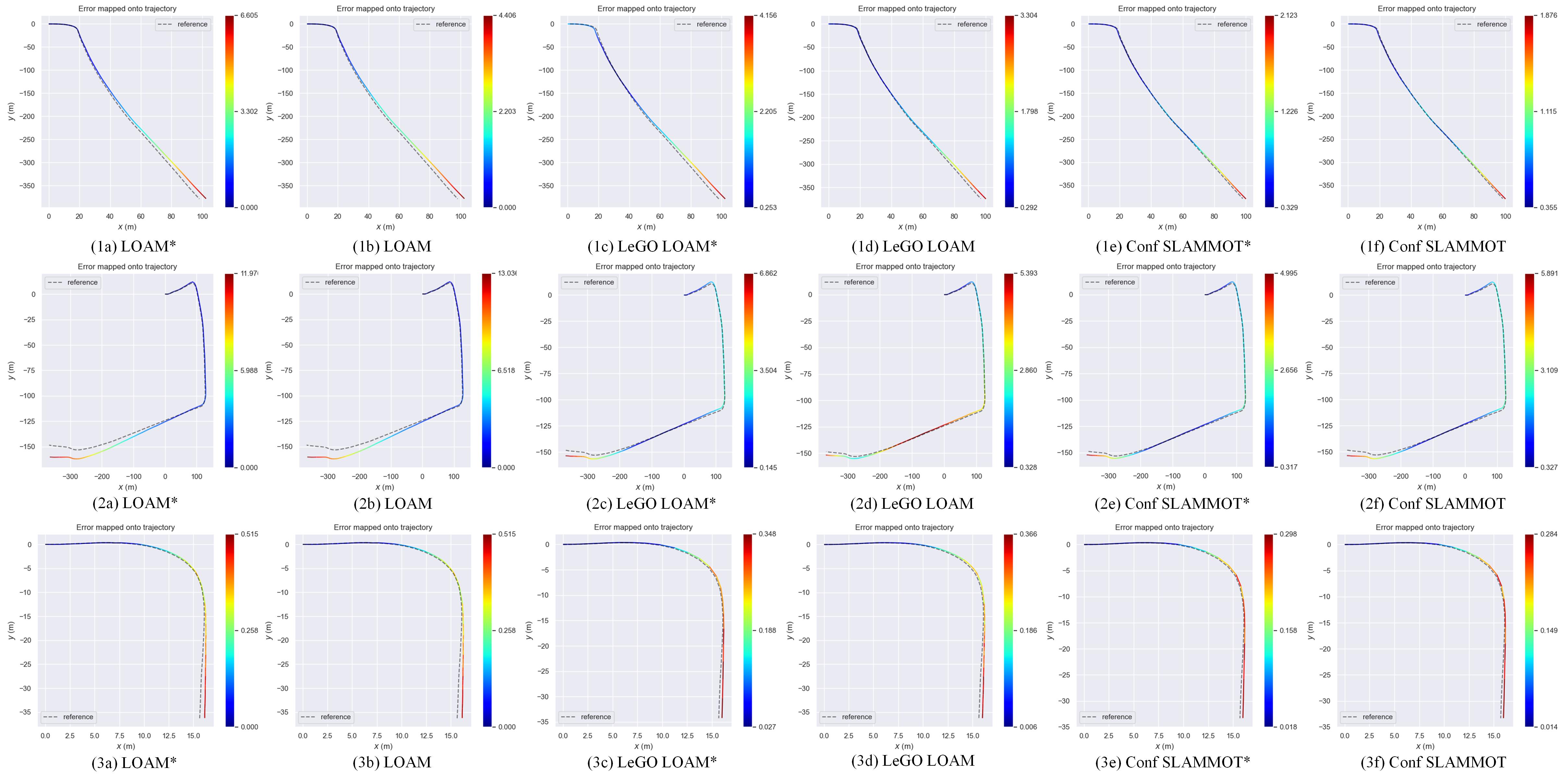}	
	\caption{Ego-trajectory error maps for different methods in KITTI Tracking dataset. The results for sequence 04, 09, 14 are shown in the first three rows. The gray dashed line denotes the ground truth.}
	\label{Fig.4}
\end{figure*}

In the ego-pose estimation evaluation experiments, we compare the pose estimation among three methods. Considering the potential different outcomes of whether to filter out dynamic object points during LiDAR odometry for scan matching, LOAM \cite{zhang2017low}, LeGO-LOAM \cite{shan2018lego}, and Conf SLAMMOT represent the methods that perform dynamic point filtering, while LOAM*, LeGO-LOAM*, and Conf SLAMMOT* indicate the methods that do not perform dynamic point filtering. We use the mean error (MEAN) and root mean square error (RMSE) as evaluation metrics. The experimental results in various sequences of the KITTI Tracking dataset are shown in Table \ref{Tab.1}. It can be observed from the table that in a few individual sequences, the performance after filtering out dynamic points is inferior to that without such filtering. This is because points belonging to potential dynamic objects (such as temporarily parked vehicles, etc.) are removed, leading to feature sparsity and thus reducing the pose estimation accuracy. For most scenarios with many moving objects and few potential dynamic objects, filtering out dynamic points to ensure the reliability of feature points can enhance pose estimation accuracy. Moreover, the pose estimation accuracy of the Conf SLAMMOT is generally higher than that of LOAM and LeGO-LOAM, indicating that in dynamic environments, tightly coupling SLAM and MOT in the graph optimization backend can benefit each other, as for SLAM, moving objects are no longer treated as interfering information but as auxiliary information to enhance the ego-pose estimation. Additionally, Fig. \ref{Fig.4} displays the ego-trajectory error maps for several sequences, providing a more intuitive comparison.
\begin{table}[htbp]
	\centering
	\caption{MEAN/RMSE of ego-pose estimation on KITTI Tracking dataset}
	\label{Tab.1}
	\begin{center}
		\resizebox{8.7cm}{1.28cm}{
			\begin{tabular}{cccccc}
				\toprule
			Methods&Seq 02&Seq 04&Seq 09&Seq 13&Seq 14\\	
				\midrule
				LOAM*\cite{zhang2017low}&1.184/1.201&2.077/2.771&4.151/5.654&0.641/0.699&0.204/0.265\\
				\midrule
				LOAM\cite{zhang2017low}&1.135/1.150&1.617/2.023&4.478/6.107&0.620/0.678&0.199/0.261\\
				\midrule
				LeGO-LOAM*\cite{shan2018lego}&1.096/1.108&1.498/1.831&2.576/3.122&0.668/0.689&0.176/0.216\\
				\midrule
			    LeGO-LOAM\cite{shan2018lego}&1.148/1.161&1.134/1.408&3.037/3.376&0.675/0.697&0.159/0.202\\
				\midrule
				Conf SLAMMOT* (Ours)&0.919/\textbf{0.923}&0.817/0.941&\textbf{1.901}/\textbf{2.288}&0.606/\textbf{0.627}&0.149/0.185\\
				\midrule
				Conf SLAMMOT (Ours)&\textbf{0.917}/0.925&\textbf{0.778}/\textbf{0.864}&2.441/2.846&\textbf{0.601}/0.632&\textbf{0.144}/\textbf{0.179}\\
				\bottomrule
		\end{tabular}}
	\end{center}
\end{table}
\subsection{Multi-object Tracking Evaluation}
We conducted several experiments on the KITTI Tracking dataset to evaluate the tracking performance of several comparative methods with different detectors, as shown in Table \ref{Tab.2}. The evaluation consider Multi-Object Tracking Accuracy (MOTA), Multi-Object Tracking Precision (MOTP), Recall and Precision based on intersection over union (IoU) thresholds of 0.5. The compared detectors include SECOND \cite{yan2018second}, PointRCNN \cite{shi2019pointrcnn} and PV-RCNN \cite{shi2020pv}, which are the three common and popular detectors with different detection quality. PV-RCNN combines the advantages of voxel-based and pointnet-based networks, reaching superior performance on various detection datasets compared to the former two \cite{shi2020pv,geiger2012we}. And the compared trackers involve AB3DMOT\cite{weng20203d}, FG-3DMOT\cite{poschmann2020factor}, PC3T \cite{wu20213d}, Conf SLAMMOT$^{\#}$ and Conf SLAMMOT. It is noted that the standalone multi-object tracking methods are evaluated by the virtual LiDAR odometry of LeGO-LOAM to provide ego-vehicle pose. Compare to Conf SLAMMOT, the Conf SLAMMOT$^{\#}$ method considers only the confidence of object predictions in the data association of tracking, without taking into account the confidence scores of object detection results. That is, the covariance matrix of the detection results during association is obtained using the overall uncertainty of the detector determined by the sensor’s error characteristics, just like in the FG-3DMOT method. From the table, it is evident that the tracking results of the PV-RCNN-based methods generally surpass those of SECOND and PointRCNN. This indirectly confirms that higher detection performance correlates with better tracking effects. Compared to Conf SLAMMOT$^{\#}$, considering the confidence scores of the detection result bounding boxes in Conf SLAMMOT will also help to adaptively adjust the implicit object association range, thereby enhancing tracking performance. From the comparative results of various tracking methods, it can be seen that the Conf SLAMMOT has achieved competitive results. The reason is that its confidence-guided implicit data association integrates the advantages of several methods, and the accuracy of simultaneously optimized ego-pose estimation by joint graph optimization backend is higher than when only using LiDAR odometry as input, which also enhances its tracking performance.
\begin{table}[htbp]
	\centering
	\caption{Evaluation result on KITTI Tracking dataset using different detectors and tracking methods}
	\label{Tab.2}
	\begin{center}
		\resizebox{8.1cm}{3.3cm}{
			\begin{tabular}{cccccc}
				\toprule
					Detector&Tracker&MOTA&MOTP&Recall&Precision\\
					\midrule
					\multirow{4}{*}{SECOND}&AB3DMOT\cite{weng20203d}&0.750&0.861&0.791&0.903\\
					\cmidrule(r){2-6}
					&FG-3DMOT\cite{poschmann2020factor}&0.779&0.872&0.859&0.915\\	
					\cmidrule(r){2-6}
					\cite{yan2018second}&PC3T\cite{wu20213d}&0.789&0.870&0.867&0.918\\
					\cmidrule(r){2-6}
					&Conf SLAMMOT$^{\#}$&0.796&0.873&0.862&0.941\\	
					\cmidrule(r){2-6}
					&Conf SLAMMOT&0.804&0.873&0.864&0.936\\	
					\midrule
					\multirow{4}{*}{PointRCNN}&AB3DMOT\cite{weng20203d}&0.809&0.865&0.847&0.904\\	
					\cmidrule(r){2-6}
					&FG-3DMOT\cite{poschmann2020factor}&0.828&0.879&0.889&0.937\\		
					\cmidrule(r){2-6}
					\cite{shi2019pointrcnn}&PC3T\cite{wu20213d}&0.831&0.881&0.895&0.934\\	
					\cmidrule(r){2-6}
					&Conf SLAMMOT$^{\#}$&0.833&0.802&0.893&0.946\\	
					\cmidrule(r){2-6}
					&Conf SLAMMOT&0.848&0.881&\textbf{0.901}&0.946\\					
					\midrule
					\multirow{4}{*}{PV-RCNN}&AB3DMOT\cite{weng20203d}&0.822&0.872&0.878&0.921\\
					\cmidrule(r){2-6}
					&FG-3DMOT\cite{poschmann2020factor}&0.837&\textbf{0.882}&0.886&0.949\\
					\cmidrule(r){2-6}
					\cite{shi2020pv}&PC3T\cite{wu20213d}&0.847&0.880&0.899&0.945\\	
					\cmidrule(r){2-6}
					&Conf SLAMMOT$^{\#}$&0.839&0.876&0.879&0.950\\					
					\cmidrule(r){2-6}
					&Conf SLAMMOT&\textbf{0.852}&\textbf{0.882}&0.898&\textbf{0.951}\\										
					\bottomrule
		\end{tabular}}
	\end{center}
\end{table}
\subsection{Overall Performance Evaluation}
We also evaluate the overall performance of Conf SLAMMOT by assessing the accuracy of state estimation for moving objects with several open-source baseline SLAMMOT methods. Specifically, we compare the multi-object tracking performance in terms of RMSE for longitudinal (long) and lateral (lat) position and yaw angle estimation, as shown in Table \ref{Tab.4}. These methods include LIO-SEGMOT \cite{lin2023asynchronous} and DL-SLOT \cite{tian2024dl},  both of which utilize LeGO-LOAM for LiDAR odometry and employ PV-RCNN for object detection to ensure consistency in comparison. Herein, several objects with different IDs are selected for error analysis. It can be seen that Conf SLAMMOT shows more advantages compared to other SLAMMOT methods. For example, the tracked frame length of object 2 in sequence 04, other methods would occur ID switching. This is because it considers the confidence of prediction and detection during tracking, which is beneficial for better data association, especially in challenging scenarios that there are continuous missed detections. The improvement in tracking results will further enhance ego-vehicle state estimation and achieve an overall performance boost through joint graph optimization.

\begin{figure*}[htbp]
	\centering
	\includegraphics[scale=1.0]{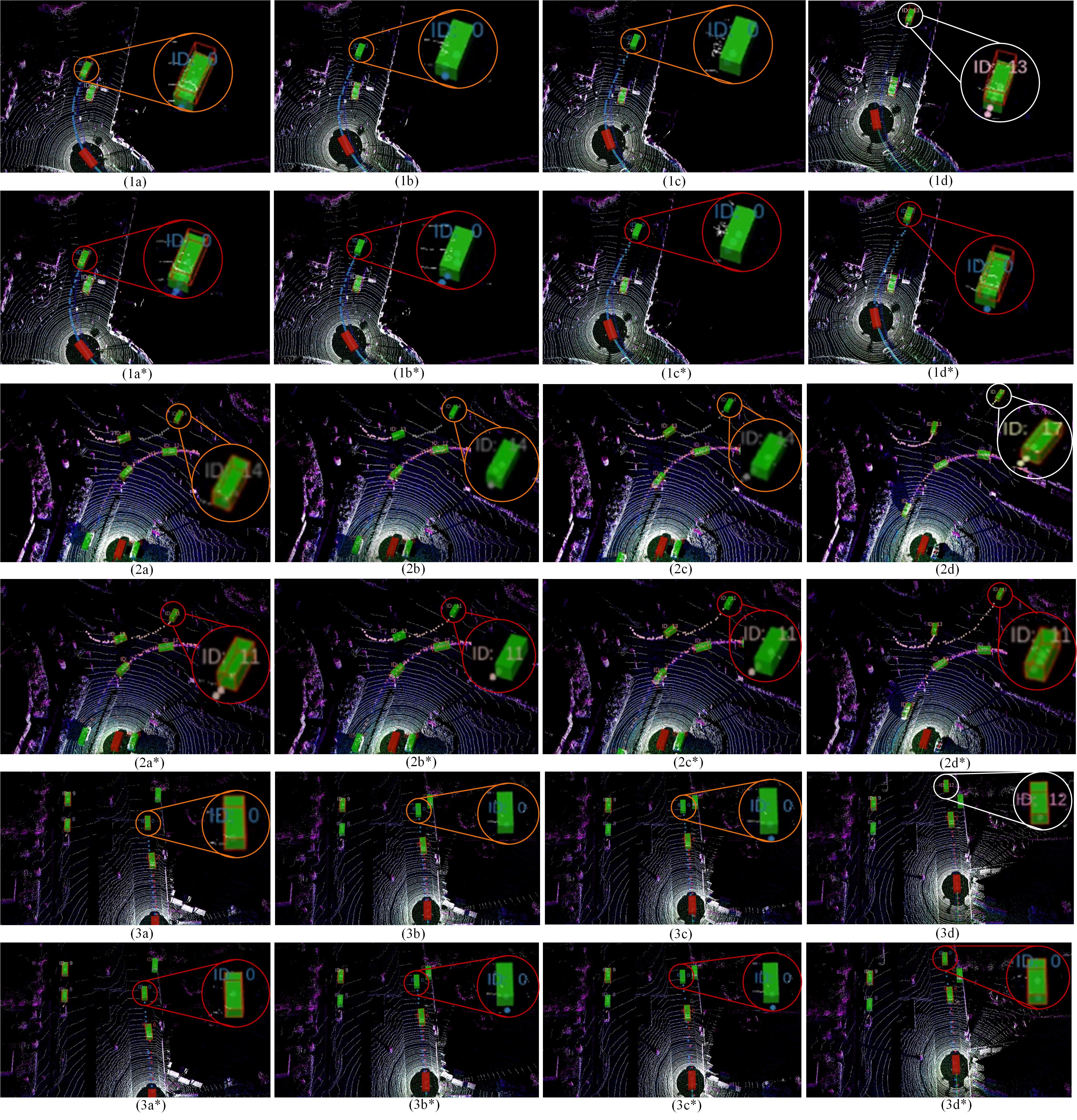}	
	\caption{The visualization results of different methods in continuous missed detection scenes. (1a-1d), (2a-2d), (3a-3d) represent the visualization results of LIO-SEGMOT at consecutive moments on sequences 00, 02, and 03, respectively; (1a*-1d*), (2a*-2d*), (3a*-3d*) represent the visualization results of the proposed Conf SLAMMOT at consecutive moments on sequences 00, 02, and 03, respectively. Red and green fully covered bounding boxes denote the ego-vehicle and tracked objects with different IDs, and red rectangular border denotes detected objects. Due to temporary occlusion or distance of objects, there are continuous frame missed detection; these objects are highlighted and enlarged with circles for display. For example, object ID 0 in (1b*-1c*), object ID 11 in (2b*-2c*), and object ID 0 in (3b*-3c*), and as shown in (1d*), (2d*), and (3d*), Conf SLAMMOT can accurately resume tracking without ID switching. In sequence 02, the reason for the differing initial object IDs between the two methods is that the vehicle has already traveled a longer distance, with more moving objects and random ID assignments. In contrast, sequences 00 and 03 represent scenarios where the vehicle has just started moving.}
	\label{Fig.5}
\end{figure*}

Furthermore, Fig. \ref{Fig.5} shows the visualization results of LIO-SEGMOT and the proposed Conf SLAMMOT method on different sequences. We tested the aforementioned compared methods on multiple sequences. Due to space limitations, we only present the results of two compared methods on representative sequences here. Additionally, considering that both LIO-SEGMOT and our proposed Conf SLAMMOT utilize factor graph-based tracking methods, the effectiveness of the confidence-guided implicit data association is more intuitively demonstrated through the figures. Three sequences (00, 02, 03) are presented, with (a-d) and (a*-d*) representing different moments of LIO-SEGMOT and Conf SLAMMOT runs. In sequence 00, object ID 1 is heavily occluded by the adjacent object for many consecutive frames, as shown in (1b-1c) and (1b*-1c*). Due to the accumulation of errors in the predicted state, ID switches can easily occur when recovering tracking, as seen in the result of LIO-SEGMOT in (1d). However, even with significant temporary occlusions, Conf SLAMMOT correctly tracked the object by successfully associating the currently detected state with the predicted state, as shown in (1d*). In sequence 02, object ID 0, as shown in (2b-2c) and (2b*-2c*), as well as in sequence 03 for object IDs 14, as shown in (3b-3c), and object ID 11, as shown in (3b*-3c*), consecutive missed detections occurred due to the objects being temporarily occluded and being far away. Compared to the ID switches that occurred in LIO-SEGMOT after recovery, as shown in (2d) and (3d), the proposed method was able to correctly track these objects, as shown in (2d*) and (3d*). These visualization results further demonstrate the effectiveness of Conf SLAMMOT. All kinds of experimental results demonstrate the superior performance of Conf SLAMMOT. It not only addresses the limitations in challenged scenes compared to other baseline SLAMMOT but also improves ego-vehicle pose estimation and multi-object tracking compared to standalone SLAM and MOT methods.
\begin{table}[h]
	\centering
	\caption{Evaluation of multi-object tracking for different methods in KITTI Tracking dataset}
	\label{Tab.4}
	\begin{center}
		\resizebox{8.6cm}{2.65cm}{
			\begin{tabular}{cccccc}
				\toprule
				\multirow{2}{*}{Methods}&\multirow{2}{*}{Seq/Object ID} & Tracked Frame&RMSE&RMSE&RMSE\\
				& &Length&(long/m)&(lat/m)&(rad)\\
				\midrule			
				\multirow{3}{*}{LIO-SEGMOT}&04/2&85&0.786&\textbf{0.791}&\textbf{0.233}\\
				\cmidrule{2-6}
				&09/15&\textbf{97}&0.339&0.982&0.217\\
				\cmidrule{2-6}
				\cite{lin2023asynchronous}&09/19&\textbf{46}&1.350&0.648&0.364\\		
				\cmidrule{2-6}
				&11/2&153&1.232&0.512&0.281\\												
				\midrule				
				\multirow{3}{*}{DL-SLOT}&04/2&89&0.679&0.810&0.278\\
				\cmidrule{2-6}
				&09/15&\textbf{97}&0.396&\textbf{0.728}&0.188\\	
				\cmidrule{2-6}
				\cite{tian2024dl}&09/19&\textbf{46}&1.301&\textbf{0.656}&0.422\\					
				\cmidrule{2-6}
				&11/2&154&1.281&0.330&0.284\\
				\midrule									
				\multirow{3}{*}{Conf SLAMMOT}&04/2&\textbf{230}&\textbf{0.653}&1.279&0.239\\
				\cmidrule{2-6}
				&09/15&\textbf{97}&\textbf{0.321}&0.778&\textbf{0.181}\\
				\cmidrule{2-6}
				(Ours)&09/19&\textbf{46}&\textbf{1.249}&\textbf{0.656}&\textbf{0.308}\\				
				\cmidrule{2-6}
				&11/2&\textbf{157}&\textbf{1.223}&\textbf{0.290}&\textbf{0.277}\\		
				\bottomrule
		\end{tabular}}
	\end{center}
\end{table}
\subsection{Implementation Efficiency Evaluation}
All of the experiments are implemented on a laptop with an Intel Core i7-10750H 2.60 GHz CPU and 12 GB RAM. And we compute the average time consumption of the main functional modules except for object detection, which can be accelerated by GPU. Specifically, the average time consumed for LiDAR odometry to process one scan is approximately 7 ms. The average time consumption for graph optimization backend (integrating the confidence-guided implicit data association) is approximately 62.4ms in the experiments. The proposed Conf SLAMMOT can be implemented  in real-time.
\section{Conclusion}
This paper proposes a Conf SLAMMOT system that tightly couples the LiDAR SLAM and confidence-guided implicit data association based multi-object tracking. This method can not only optimize the estimated state of both the ego-vehicle and moving objects simultaneously, but also effectively recovers tracking of continuously missed detections based on confidence-guided data association, leading to a stable backend and improved estimation performance. Comparing Conf SLAMMOT with other baseline methods, various experiments demonstrate that the presented method achieves competitive accuracy and performance in terms of ego-pose estimation and object state estimation. In addition, our experiments demonstrate the superior performance of Conf SLAMMOT in scenarios where there are some missed detections due to objects being either occluded or distant, which is meaningful in real-world applications.

Moreover, in this work, we neglect objects or the surrounding environment with different categories, which may be detrimental not only to the proposed Conf SLAMMOT method but also to related object perception methods. Therefore, incorporating semantic information will be a direction for future work. Additionally, integrating multiple motion models into multi-object tracking for dealing with moving objects with ambiguous motion status and irregular motion patterns may be a promising extension of research works.
\ifCLASSOPTIONcaptionsoff
\newpage
\fi

\bibliographystyle{IEEEtran}
\bibliography{Refs_Confidence_Index}
\begin{IEEEbiography}[{\includegraphics[width=1in,height=1.25in,clip,keepaspectratio]{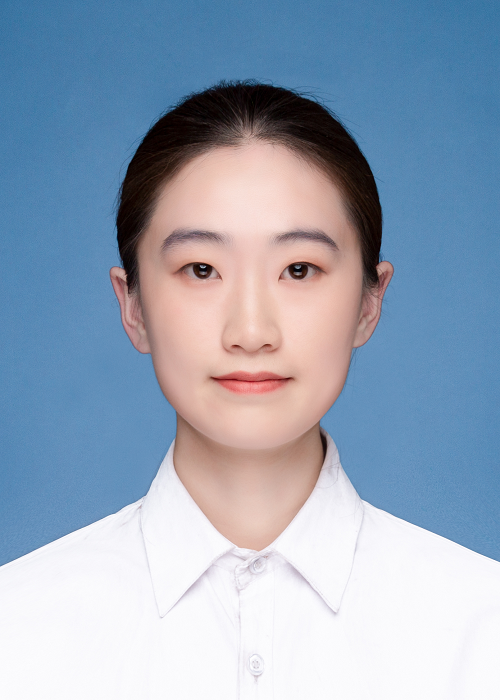}}]{Susu Fang},
currently a Ph.D. candidate at the Department of Automation, Shanghai Jiao Tong University, Shanghai, China. She obtained the M.Eng. degree and B.Eng. degree from Shandong University and Nanjing Agricultural University, China, in 2019 and 2016 respectively. Her research interests are in autonomous vehicle localization and perception, multi-sensor data fusion, and multi-vehicle cooperative intelligent systems.
\end{IEEEbiography}
\newpage

\begin{IEEEbiography}[{\includegraphics[width=1in,height=1.25in,clip,keepaspectratio]{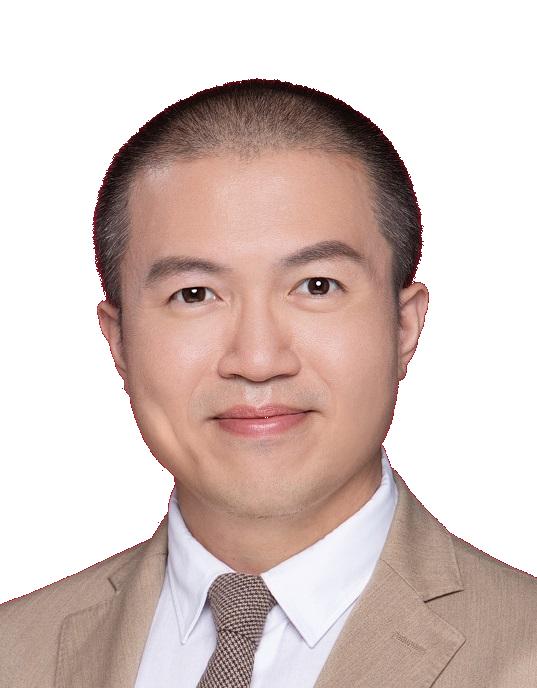}}]{Hao Li}, 
associate professor and doctoral supervisor with SPEIT and the Department of Automation, Shanghai Jiao Tong University (SJTU), China. He obtained the Ph.D. degree from the Robotics Center of MINES ParisTech and INRIA in 2012, the M.Eng. degree and B.Eng. degree from the Department of Automation of SJTU in 2009 and 2006 respectively. His current research interests are in automation, computer vision, data fusion, and cooperative intelligent systems.
\end{IEEEbiography}
\end{document}